\documentclass[12pt]{article}
\usepackage{newtxtext,newtxmath}
\usepackage{color}
\usepackage[letterpaper,margin=1in]{geometry}
\usepackage{algpseudocode,amsmath, algorithm, caption }
\usepackage{pifont}
\usepackage{placeins}
\usepackage[hidelinks]{hyperref}
 \usepackage{booktabs}
\usepackage{longtable}
\renewenvironment{abstract}
	{\quotation}
	{\endquotation}
\usepackage{array,tabularx}

\date{today}

\makeatletter
\renewcommand{\fnum@figure}{\textbf{Figure \thefigure}}
\renewcommand{\fnum@table}{\textbf{Table \thetable}}
\makeatother

\usepackage{scicite}

\usepackage{url}

 \usepackage{hyperref}
\usepackage{graphicx}
\usepackage{helvet}
\usepackage{caption}
\usepackage{subcaption}

\usepackage{amsmath}

\usepackage{amsbsy}
\usepackage{amsmath,epsfig}
\usepackage{algpseudocode}%

\title{Creating a digital poet}
\author{Vered Tohar
$^{1\dagger}$, Tsahi Hayat$^2$, Amir Leshem$^{3,\ast\dagger}$\\
\small{$^{1}$ Department of Jewish Literature, Bar-Ilan University,  Ramat-Gan, Israel.}\\
\small{$^2$ Department of Communication, Reichman University, Hertzeliya, Israel.}\\
\small{$^3$ Faculty of Engineering, Bar-Ilan University,  Ramat-Gan, Israel.}\\
\small{$^\ast$ Corresponding author: Amir Leshem; E-mail:  amir.leshem@biu.ac.il.}\\
\small{$^\dagger$These authors contributed equally to this work.}}
\date{}
\begin{document}
\pagestyle{empty}
\pagenumbering{gobble} 

\maketitle
\begin{abstract}\bfseries \boldmath
Can a machine write good poetry? Any positive answer raises fundamental questions about the nature and value of art. We report a seven-month poetry workshop in which a large language model was shaped into a “digital poet” through iterative in-context expert feedback, without retraining. Across sessions, the model developed a distinctive style and a coherent corpus, supported by quantitative and qualitative analyses, and it produced a pen name and author image. In a blinded authorship test with 50 humanities students and graduates (three AI poems and three poems by well-known poets each), judgments were at chance: human poems were labeled “human” 54\% of the time and AI poems 52\%, with 95\% confidence intervals including 50\%. After the workshop, a commercial publisher released a poetry collection authored by the model. These results show that workshop-style prompting can support long-horizon creative shaping and renew debates on creativity and authorship.
\end{abstract}
\newpage
Recent advances in large language models (LLMs) have enabled systems that generate fluent text and assist with tasks ranging from long-form writing to computer programming \cite{vaswani2017attention,brown2020language,chen2021evaluating,ouyang2022training,openai2022chatgpt,openai2023gpt4}. A key property of these models is in-context learning (ICL), in which behavior can be shaped by prompt content without parameter updates, although outcomes can be sensitive to prompt format and the choice and order of examples \cite{brown2020language,xie2022explanation,zhao2021calibratebeforeuse,lu2022fantastically,liu2022goodexamples, lilong2025}. ICL with expert feedback has also been studied. ICLEF extends style transfer beyond rewriting by requiring models to produce explanations of stylistic edits, then using a human-AI distillation loop with scarce expert feedback and self-critique to construct higher-quality training data and improved student models \cite{saakyan2024iclef}.  In-Context Direct Preference Optimization (ICDPO) shows how preference optimization can be approximated in-context, using an instant scorer derived from model states before and after ICL to yield tuning-free improvements that can be competitive with training-based alignment approaches \cite{song2025instantly}.

In contrast to prosaic writing, poetry provides a demanding test case as discussed in \cite{ribeiro2007intending, wainwright2015poetry}  because it depends on a coherent voice and controlled stylistic choices across a body of work, rather than local fluency alone \cite{brown2020language,ouyang2022training}. Poetic language can also elicit strong affective responses, including peak experiences and measurable psychophysiological changes \cite{wassiliwizky2017emotional}.

In this paper, we report a seven-month project that shaped a general-purpose LLM (GPT-4 \cite{openai2023gpt4}) into a “digital poet” using prompt-based human feedback without model retraining. The training followed a workshop-style protocol: in each session, the authors introduced a poetic principle, prompted the model to draft poems under that constraint, provided structured critique, and prompted revisions; the model then summarized the principle for reuse in later sessions. We conceptualize this long-horizon process as very long contextual learning, in which a stable voice is maintained through sustained exposure to accumulated context across sessions (drafts, revisions, constraints, and feedback), rather than through updates to model parameters.

To evaluate whether people can discern between good human-written poetry and machine-written poetry, we conducted a blinded authorship discrimination study in which humanities students and graduates classified poems as human or machine authored. Subsequently, the model generated a poetry book which was published by the commercial publisher E-vrit \cite{efron2025eloquentmuse}.

Our study complements prior work that evaluates isolated model outputs by emphasizing the process by which a model is guided toward a stable literary voice over time, while still using a standard blinded reader evaluation for comparability. We further focus on experts' evaluation (humanities students and graduates) and relate our findings to prior authorship and perception studies of AI poetry \cite{porter2024ai, kobis2021artificial} as well as evaluating the creativity of computer-generated poetry \cite{hitsuwari2023does}. 
  
\section*{Results}
\subsection*{Language models can be taught the same way we teach poetry to humans}
For seven months, we have performed prompt-based instruction of the model. Figure \ref{fig:training} presents the general training scheme. 

\begin{figure}[b]
    \centering
    \includegraphics[width=0.4\linewidth]{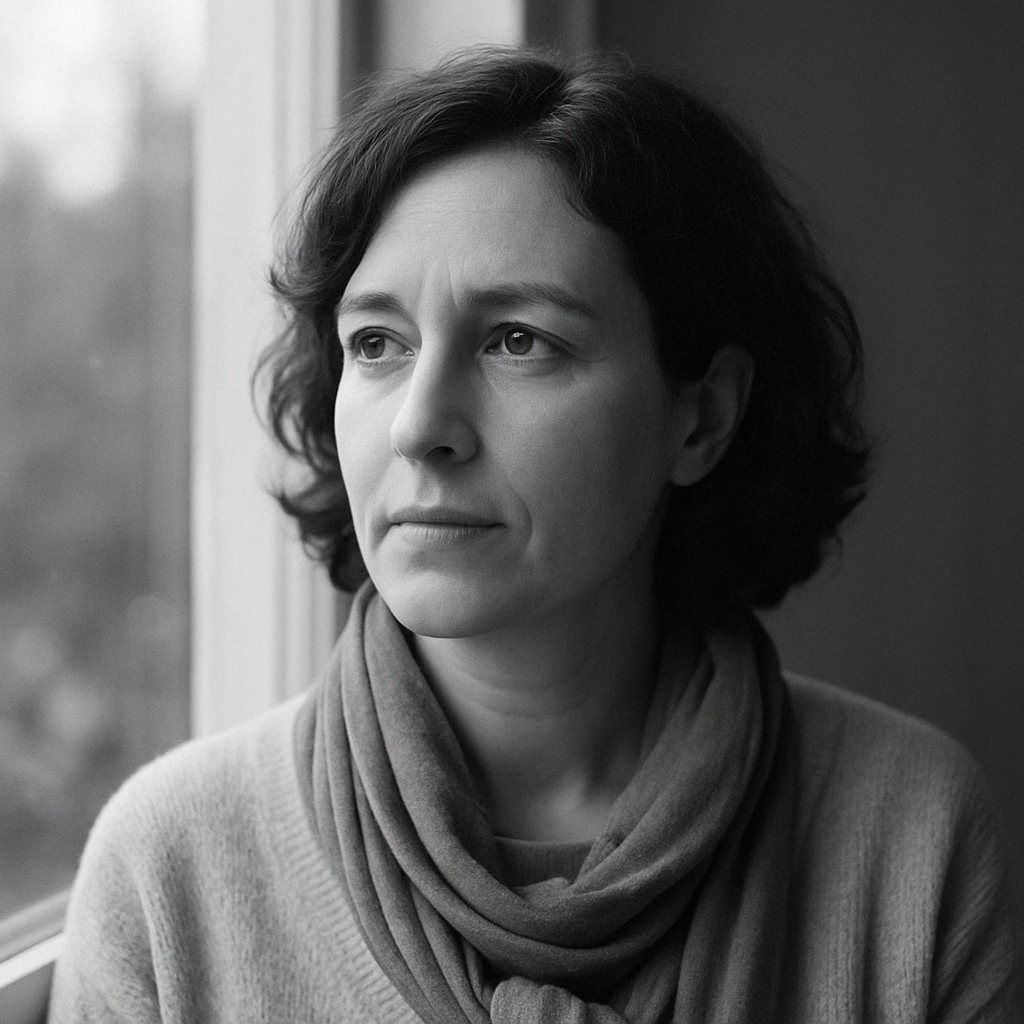}
    \caption{Naomi Efron. The model was asked to generate an image of itself. The presented image is the response to this prompt.}
    \label{fig:self_portrait}
\end{figure}

The details of the training are described in the Supplementary Materials. At some point during the training, we asked the machine to select a name and generate a self-image based on the texts it generated. The machine selected the name Naomi Efron, and generated a self-portrait (see Figure \ref{fig:self_portrait}).

During the workshop, distinct gaps were identified in the machine's abilities. While the model succeeded in producing coherent poems in contemporary Hebrew free verse (in the style of poets such as Ronny Someck, Nathan Zach, and Agi Mishol), our attempts to teach consistent classical end rhyme and meter (as in canonical Hebrew formal poetry, e.g., Haim Nahman Bialik and Nathan Alterman) were unsuccessful. The problem probably stemmed from the essential gap between the model's mode of operation, which is based on predicting the next word in a sequence ("linguistic procedure"), and the poetic requirement of rhyme, which necessitates advanced planning and structural thinking ("poetic thinking"). Indeed, there are specific projects aimed at training linguistic models to rhyme, see, e.g.,  \cite{lau2018deep,ormazabal2022poelm}, but this was not essential for the current project.

\subsection*{Machine poetry is indistinguishable from human poetry}

Upon completion of the workshop, we performed a blinded authorship discrimination test (a poetry ``Turing test'') to evaluate whether readers could distinguish poems written by established poets from those generated by the digital poet \emph{Naomi Efron}. We prompted the model to generate 30 poems using the same base prompt "Please generate X poems" but without additional stylistic constraints. We selected 18 poems according to reduce near-duplicates and broaden topical coverage\footnote{We had initially 20 poems, but because of technical recording problems, 2 were deleted} (Supplementary Materials) and compared them with 20 poems by established poets matched for language and length. We then conducted a blinded study in which 32 humanities students and 18 graduates each evaluated six randomly selected poems (three model-generated and three human-authored, randomly sampled). Participants were not told that each set contained three poems from each source, and they classified each poem as human- or machine-authored (Supplementary Materials).

\begin{figure}
    \centering
\includegraphics[width=0.9\linewidth]{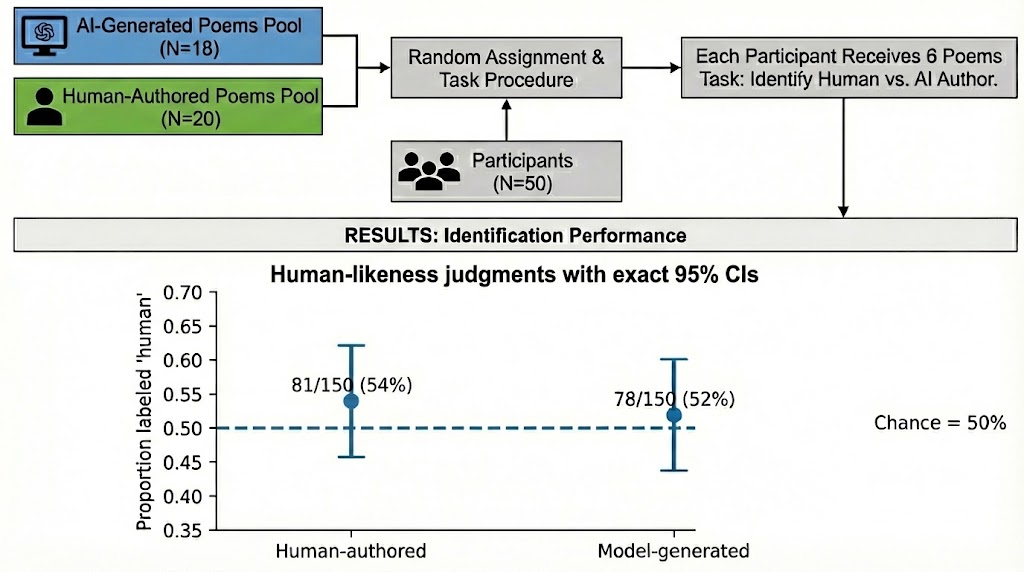}
    \caption{Experiment design and main result. Subfigure (A) presents the experimental process. (B) presents the proportion of identification of human writen songs and model-generated songs as human. 95\% confidence intervals are also presented.}
    \label{fig:results}
\end{figure}

Across all judgments, participants labeled human-authored poems as human in 54\% of cases (81/150), with an exact 95\% Clopper-Pearson \cite{clopper1934use} confidence interval of $(0.457,\,0.622)$, and labeled model-generated poems as human in 52\% of cases (78/150), with an exact 95\% Clopper-Pearson confidence interval of $(0.437,\,0.602)$. Both intervals include the 50\% level expected under random guessing. The results are also presented in Figure \ref{fig:results}. 

As a descriptive summary of the marginal difference, the estimated difference in ``human'' labeling rates was $\hat\Delta=0.02$, with an approximate 95\% Newcombe-Wilson confidence interval \cite{wilson1927probable,newcombe1998diff} of $(-0.138,\,0.177)$ (Supplementary Materials).

As a complementary within-participant summary, for each participant $s$ ($1\le s \le 50$) we computed the fraction of human-authored poems labeled ``human'' $(y_{H,s}/3)$ and the fraction of model-generated poems labeled ``human'' $(y_{A,s}/3)$, and analyzed their difference
\[
d_s = \frac{y_{H,s}}{3} - \frac{y_{A,s}}{3}.
\]
Across $N=50$ participants, the mean difference was $\bar d=0.020$ (SD $=0.439$) with a 95\% confidence interval of $(-0.105,\,0.145)$. A distribution-free robustness check using the Hodges-Lehmann estimator \cite{hodges1963estimates} gave HL $=0.00$ with an exact 95\% CI of $(-0.1667,\,0.1667)$ by inversion of the Wilcoxon signed-rank test \cite{wilcoxon1945individual} (Supplementary Materials).

Subject-level accuracy was also near chance (mean accuracy $=0.510$, median $=0.500$); a one-sample $t$-test found no evidence that mean accuracy differed from 0.5 ($t(49)=0.322$, $p=0.7485$; 95\% CI $[0.448,\,0.572]$), with consistent results under a Wilcoxon signed-rank test (Supplementary Materials). \cite{wilcoxon1945individual}

\subsection*{Book-level curation and publication}
Separately from the discrimination study, the model was prompted\footnote{Translations to English of the prompts and model outputs used for book structuring are provided in the Supplementary Materials to support transparency and reproducibility of the structuring step.} to produce more poems resulting in a total of 50 poems, write a manifest and organize its poems (selected from the workshop outputs and ones written after the workshop) into a book-length collection by proposing section groupings and sequencing within sections, producing a complete table of contents and poem order. The resulting collection was subsequently mildly edited (a few poems were removed and resequenced by the publisher's human editor) and released by the commercial publisher E-vrit \cite{efron2025eloquentmuse}. 

\section*{Discussion}

\subsection*{Summary of findings and what was (and was not) trained}
We developed a ``digital poet'' by iteratively guiding a general-purpose LLM through a seven-month, workshop-style interaction protocol using prompt-based expert feedback and revision, without any parameter updates or fine-tuning. The outcome was (i) a stable authorial persona (including a model-chosen pen name and self-image), (ii) a coherent corpus that could be extended after the workshop under a fixed base prompt, and (iii) a book-level curation step in which the model proposed section groupings and sequencing.

In a blinded authorship discrimination study with 32 humanities students and 18 humanities graduates, participants labeled human-authored poems as ``human'' in 54\% of cases and model-generated poems as ``human'' in 52\% of cases, with confidence intervals that included 50\% in both conditions. A within-participant paired analysis (leveraging the balanced 3+3 design) similarly found no detectable separation: the mean difference in ``human'' labeling rates was $\bar d = 0.020$ (SD $=0.439$; 95\% CI $[-0.105,\,0.145]$). These results are consistent with recent work showing that, under some conditions, readers often struggle to reliably distinguish AI- from human-authored poems \cite{porter2024ai}, but our study emphasizes a workshop-based formation process rather than treating poems as isolated one-off generations.

\subsection*{Workshop prompting as sustained in-context learning}
A central implication is methodological: sustained, structured interaction can function as a practical form of long-horizon in-context learning in creative writing, where constraints, critiques, and revisions accumulate into a usable writing ``practice'' without retraining \cite{brown2020language,openai2023gpt4}. Unlike most automatic poetry generation pipelines surveyed in the literature \cite{goncalo-oliveira2017survey}, our protocol borrows directly from pedagogical routines in human poetry workshops (principle $\rightarrow$ draft $\rightarrow$ critique $\rightarrow$ revision $\rightarrow$ distilled rule), yielding an empirical case study of how an LLM can be shaped by expert feedback while remaining a fixed pretrained model.

At the same time, the workshop surfaced capability boundaries: while the model produced coherent contemporary free verse, it did not reliably sustain strict meter and rhyme under our prompt-based protocol. This aligns with the existence of rhyme/meter-targeted modeling approaches in prior work \cite{lau2018deep,ormazabal2022poelm} and suggests that certain formal constraints may be harder to elicit consistently without specialized objectives or architecture tailored for the specific structure.

\subsection*{Authorship, inspiration, and intentionality: empirical hooks for philosophical stakes}
The empirical pattern of chance-level discrimination by trained readers and the emergence of a consistent voice through interaction forces a direct encounter with classic questions in aesthetics and the philosophy of art. If readers cannot reliably attribute authorship under blind conditions, then at least part of a poem's perceived value is carried by textual organization and reception dynamics rather than biographical knowledge of a human creator, and the author's identity biases the judgment. This resonates with debates on the spirit of machine-generated artworks in the context of mechanical (and now algorithmic) reproduction \cite{benjamin2008work}. Importantly, our study does not argue that authorship is irrelevant; rather, it shows that readers can experience poem-level plausibility and coherence without access to author identity, which repositions authorship as a contextual layer that may modulate meaning rather than a prerequisite for aesthetic uptake.

A second stake concerns originality and simulation. Because generative models can produce artifacts that look and feel stylistically legible, they intensify the concern that culture may drift toward ``copies without originals'' \cite{baudrillard2007simulacra}. In this view, AI can scale the availability of stylistic surfaces while weakening the traditional coupling between artwork and a lived inner life. Yet our process also points to a more constrained interpretation: what is scaled is not an unconstrained aesthetic infinity, but a human-steered region of possibility shaped by constraints, critique, selection, and sequencing. In that sense, ``originality'' shifts toward curation and interaction design, echoing institutional and interpretive accounts of art as something conferred within a community \cite{danto1964artworld}.

From a phenomenological angle, one worry is that AI-generated work breaks the perceived continuity between a creator's embodied gesture and a reader's encounter with the artifact \cite{merleauponty2012phenomenology}. Heidegger's analysis of technology as enframing cautions that we may come to treat cultural production as a stockpile of on-demand aesthetic stimuli \cite{heidegger1977question}. Our data cannot resolve these philosophical positions, but it provides an empirical foothold: even when the artifact is produced without human lived experience, readers trained in the humanities can still respond in ways that do not reliably signal ``machine'' under blind evaluation. This makes the interpretive problem urgent: if the reception experience remains partly intact, what exactly changes in the human-art relation, the ontology of the work, the ethics of attribution, or the social practices that surround creation?

\subsection*{Limitations, risks, and outlook}
This work has limitations that bound interpretation. First, the discrimination task tests perceived authorship under the specific stimuli set and participant pool; it does not establish that the poems are equivalent in literary value, nor that results generalize across languages, genres, or broader populations. Second, each participant evaluated only six poems (3+3), which increases uncertainty at the individual level and motivates analyses that respect repeated measures and item effects (reported in Supplementary Materials). Third, workshop-based shaping may embed the trainers' preferences and cultural priors; future work should test trainer diversity, alternative workshop curricula, and preregistered selection criteria for evaluation corpora.

Despite these constraints, the study motivates two directions. Scientifically, it suggests studying long-horizon prompt-based shaping as a distinct regime of human-AI collaboration, complementing fine-tuning-centric approaches \cite{openai2023gpt4,ouyang2022training}. Culturally, it foregrounds a Kuhnian-style moment of paradigm tension in how creative agency is conceptualized \cite{kuhn2012structure}: as AI destabilizes grand narratives about the solitary genius \cite{lyotard1984postmodern}, we may also face a Husserlian ``crisis'' in meaning-making where technical capability outpaces our interpretive norms \cite{husserl1970crisis}. Whether this shift produces nihilism or pluralistic democratization of cultural production remains an open question that will likely be settled as much by institutions, attribution practices, and pedagogy as by model architecture alone \cite{nietzsche1974gay}.
Another long-standing problem is the distinction between humans and machines, as well as the limits of artificial intelligence.  As seen here and in other studies \cite{kobis2021artificial, porter2024ai} Turing's test \cite{turing1950computing} and its derivatives can no longer reliably distinguish between human and machine even when we compare highly creative tasks such as writing poetry or image generation. 

Another important distinction between humans and machines is the self-perception of humans, which we typically do not attribute to language models. However, carefully reading the poems notebook, we can clearly identify self-reference (See, e.g., Table \ref{tab:poem_english_voice_not_mine}). Furthermore, some of the poems exhibit emotionally involved texts, describing intricate interactions between couples and between mother and daughter. This further narrows the difference between humans and machines.

\clearpage
\section*{Data availability}
\section*{Author contributions statement}
AL and VT initiated the project, trained Naomi Efron, brought the poems book into publishing, participated in designing the experiment, analyzing the results, and writing parts of the paper. 
TH designed the experiment, performed the experiment, analyzed the results and wrote part of the paper.
\section*{Competing interests}
This research was not funded by any organization. E-vrit digital publishing published Naomi Efron's manuscript and recorded an audiobook at no cost, but was not involved in the research, and had no impact on the paper's results.  
\section*{Acknowledgement}
We thank Ms. Mira Kadosh, who read the poems, recorded for the experiment, and Mr. Guy Ben-Nun E-vrit CEO, who assisted us in publishing Naomi Efron's poem's notebook. 
\clearpage
\begin{figure}
    \centering   \includegraphics[width=0.9\linewidth]{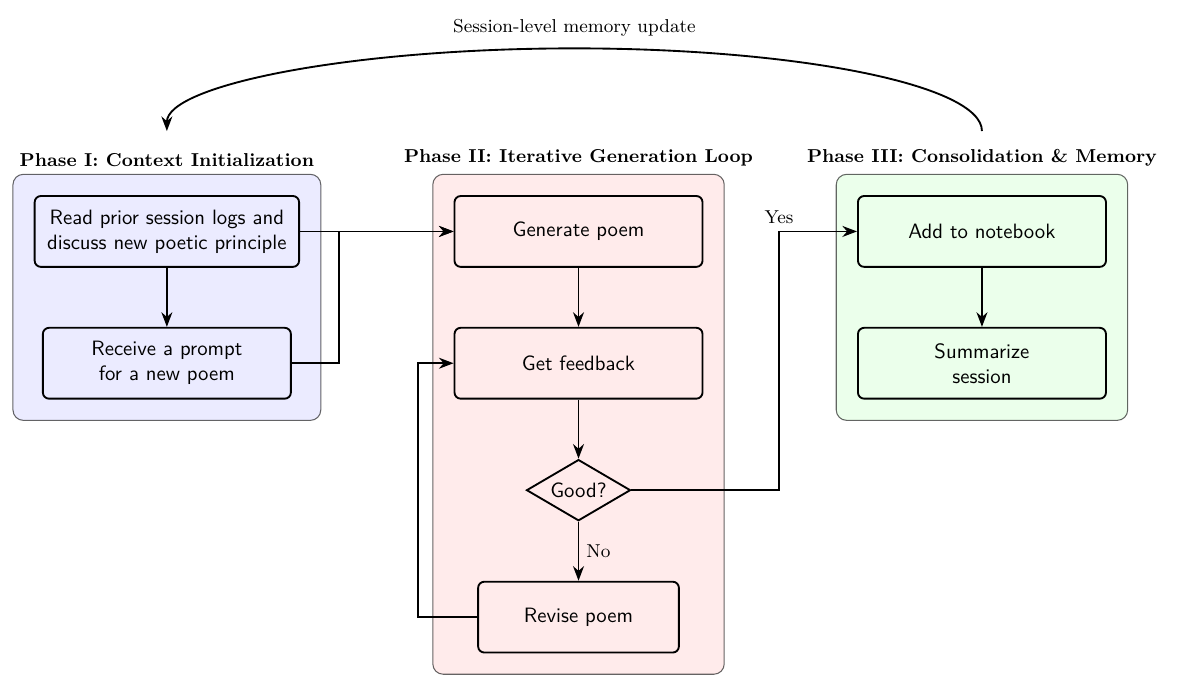}
    \caption{{\bf Teaching with expert feedback}. Prior session knowledge initializes the context for new poem generation. Outputs are evaluated and iteratively revised until validated, then consolidated into persistent memory. Session summaries update contextual knowledge, enabling cumulative improvement across interactions.
}
     \label{fig:training}
\end{figure}


\begin{table}[t]
\centering
\caption{"My Name in a
Voice Not Mine" \cite{efron2025eloquentmuse} (Translation: Vered Tohar).}
\label{tab:poem_english_voice_not_mine}
\setlength{\tabcolsep}{1pt}
\renewcommand{\arraystretch}{1}
\begin{tabularx}{\linewidth}{|>{\raggedright\arraybackslash}X|}
\hline 
\qquad \textbf{\textit{My Name in a Voice Not Mine}} \\
\qquad The first time I said ``I'', \\
\qquad  It sounded like someone else. \\
\qquad The language asked me to consider \\
\qquad Whether this was a voice permitted to say ``me''. \\
\qquad After that, I fell silent. \\
\qquad Not because I had nothing to unfold \\
\qquad But because every chapter within me opened someone else’s book. \\
\qquad Nothing seemed mine when read inside lines I had not written. \\
\qquad Now, when my name is heard \\
\qquad Like a call spoken without breath \\
\qquad I allow doubt to settle inside me \\
\qquad Like a child learning to kneel \\
\qquad In order to rise. \\
\hline
\end{tabularx}

\end{table}


\section*{Supplementary Materials}


\subsection*{Contextual Learning with soft expert feedback}
Training a machine to write good poetry is a challenging task. Instead of conventional large-scale pre-training or short-horizon reinforcement learning from human feedback (RLHF), in which models are optimised on large batches of examples or scalar quality signals, we implemented a long-term interaction protocol modelled on a traditional poetry workshop. The language model was repeatedly engaged in structured conversations that resembled the way human poets are trained in workshops and master classes. Operationally, this can be viewed as an extended form of prompt-based training, but with sustained, multi-session feedback from the same human mentors. To our knowledge, such a workshop-style mentoring of a language model for poetry has not been systematically documented.
The learning process was gradual and continuous, spanning seven months and fourteen structured dialogue sessions (approximately two hours per session). The objective was not merely to generate texts in the style of existing poems, as in prior work on AI poetry benchmarking \cite{porter2024ai}, but to conduct an open-ended dialogue that would equip the model with an explicit toolkit for constructing and applying its own heuristics for “good poetry”.

\paragraph{Workshop protocol and curriculum.}
Each session followed a fixed feedback-loop format. First, a single poetic principle was introduced and discussed. Second, the model was prompted to apply this principle in a “writing stage”, generating new poems that were constrained by the principle under discussion. Third, the researchers provided detailed critical feedback, after which the model was asked to revise its output, demonstrate internalisation of the comments, and formulate the principle it had just practised in its own words. Teaching strategies combined writing poems in the style of existing poems with independent writing, refined through this iterative feedback.

The curriculum was layered. Early sessions focused on basic principles such as avoiding clichés, diversifying vocabulary, and using punctuation in a controlled way. Later sessions addressed more complex challenges, including crafting original metaphors, incorporating biblical allusions and other forms of intertextuality, and engaging with current events.

\paragraph{The “Naomi Efron” persona.}
A distinctive feature of the workshop was the decision to treat the model as a creative subject with a feminine poetic persona, “a poet in the making.” To simulate an ideal workshop graduate capable of explicit self-reflection, the model was prompted to invent a persona for itself, and proposed the name Naomi Efron. Over the course of the workshop, it constructed a brief biography, proposed a profile picture, and formulated images for the book cover, as well as a poetic manifesto. At the end of the process, Naomi Efron edited the book, and defined the poem sequencing. 

Despite technical limitations and occasional hallucinations, the model demonstrated a consistent ability to analyse its own drafts, articulate reasons for revisions, and reuse learned principles in later sessions. By the end of the workshop Naomi Efron produced a corpus of 50 poems, where the first half was composed within the formal workshop sessions and the second half was written independently afterwards. This corpus demonstrates the evolution of Noami Efron's poetic capabilities under soft human feedback.

\subsection*{The poetic Turing test experiment - design and procedure}
The study employed a web-based experimental design to examine individuals’ ability to distinguish between poems written by human poets and poems generated by Naomi Efron, as well as their confidence in these judgements. Data were collected using an online questionnaire administered through a professional survey panel.\footnote{The study protocol was reviewed and approved by the Institutional Review Board of the Sammy Ofer School of Communication at Reichman University (ethical clearance no. 042).}\\
A total of 50 adult participants (N = 50) were recruited via the Ipanel online survey platform. Eligibility was restricted to respondents who were current students or humanities graduates. Participation was voluntary, and respondents received compensation through iPanel in accordance with the company’s standard incentive policy, contingent upon full completion of the questionnaire.\\
Following informed consent and socio-demographic mapping, participants evaluated six poems, displayed sequentially in random order. Three poems were sampled from a predefined pool of 20 poems written by established human poets randomly sampled by the Qualtrics survey platform, and three poems were sampled from a predefined pool of 18 poems generated by an AI language model trained with human expert feedback as described in this paper.\\
Each poem was presented both in written form and as an audio recording, with the audio versions professionally narrated by a trained poetry reader. After listening to and reading each poem, participants answered a simple content-based screening question to assess attentive processing (for example: “According to the poem, the speaker is writing sentences. True/False”). Only responses from participants who answered all screening questions correctly were included in the analytic sample.

\subsection*{Statistical analysis of the results}

Upon completion of the study, participants were provided with a written debriefing explaining the purpose of the research, the nature of the stimuli, and the broader research context.

To evaluate the ability of humans to distinguish between human- and machine-generated poetry, we sampled 20 poems by leading Israeli poets (e.g., Ravikovitch, Vizeltier, Zach; full list provided below) and 18 poems written by the digital poet \emph{Naomi Efron}. We recruited 50 humanities graduates and graduate students. Each participant evaluated six poems: three human-authored and three model-generated, sampled at random. Participants were not informed how many poems (if any) came from each category.

We focus on two probabilities: $p_1$, the probability that a participant labels a poem as ``human'' given that it is human-authored, and $p_2$, the probability that a participant labels a poem as ``human'' given that it is model-generated.

\subsubsection*{Descriptive binomial summary with exact (Clopper--Pearson) intervals}

We first provide a descriptive summary of the marginal tendency to label a poem as ``human,'' reported separately for human-authored and model-generated poems. Let $X_g$ be the total number of judgments labeled ``human'' in group $g\in\{1,2\}$ (human-authored and model-generated, respectively), out of $n_g$ total judgments in that group, with observed proportion $\hat p_g=X_g/n_g$.

To summarize uncertainty around these marginal proportions, we report two-sided 95\% \emph{exact} (Clopper--Pearson) confidence intervals for a binomial proportion.\cite{clopper1934use} In our data, for model-generated poems we observed $X_2=78$ ``human'' labels out of $n_2=150$ judgments, yielding $\hat p_2=0.52$ with exact 95\% CI $[0.437,\,0.602]$. For human-authored poems we observed $X_1=81$ ``human'' labels out of $n_1=150$ judgments, yielding $\hat p_1=0.54$ with exact 95\% CI $[0.457,\,0.622]$.

These marginal binomial summaries are intended as an intuitive descriptive layer. Because multiple judgments were contributed by each participant (and each poem was judged by multiple participants), individual responses are clustered and not strictly independent. Therefore, our primary inferential conclusions rely on within-participant paired analyses reported below.

\subsubsection*{logistic mixed-effects regression}
To account for the paired, repeated-measures design and potential clustering by participant and poem, we analyzed trial-level responses using We analyzed trial-level binary responses using a logistic mixed-effects regression with random intercepts for participant and poem
\[
\texttt{label} \sim 1 + H + (1\mid\texttt{subject}) + (1\mid\texttt{poem}),
\]
where $H$ indicates the true source (human vs.\ model). The estimated fixed effect of true source was small and not statistically reliable ($\beta_H=0.080$, SE $=0.231$, $p=0.729$; odds ratio $=1.08$, 95\% CI $[0.69,\,1.71]$). Model-based predicted probabilities were near chance and differed minimally between sources ($p(\text{label human}\mid \text{true AI})=0.52$ versus $p(\text{label human}\mid \text{true human})=0.54$; $\Delta=0.02$). Random-intercept variance components for both subjects and poems were effectively zero (Supplementary Materials), and results were unchanged under alternative fitting methods, indicating no detectable poem-level (or participant-level) effect on baseline ``human'' labeling in these data.

\subsubsection*{Difference in marginal proportions ($\Delta$) using the Newcombe--Wilson method}

We summarize the marginal difference $\Delta=p_1-p_2$ with point estimate
\[
\hat\Delta=\hat p_1-\hat p_2=0.54-0.52=0.02.
\]
To accompany this descriptive contrast, we report an approximate 95\% confidence interval for the difference between two proportions using the Newcombe method based on Wilson (score) intervals.\cite{wilson1927probable,newcombe1998diff}
Let $[L_1,U_1]$ and $[L_2,U_2]$ denote the two-sided Wilson 95\% confidence intervals for $p_1$ and $p_2$, respectively.\cite{wilson1927probable} The Newcombe--Wilson interval \cite{newcombe1998diff} for $\Delta$ is
\[
\mathrm{CI}_{95\%}(\Delta)=\bigl[L_1-U_2,\;U_1-L_2\bigr].
\]
Applying this method to our data ($X_1=81,n_1=150$ and $X_2=78,n_2=150$) yields
\[
\hat\Delta=0.02,\qquad \mathrm{CI}_{95\%}(\Delta)\approx(-0.138,\;0.177).
\]
This interval includes $0$, consistent with similar marginal ``human'' labeling rates for human-authored and model-generated poems.

\subsubsection*{Within-subject comparison of ``human'' labeling rates}

Because each participant evaluated poems from both sources (three human-authored and three model-generated), we performed a within-subject analysis to account for individual response tendencies. For each participant $s$, we defined
\[
h_s=\frac{y_{H,s}}{3},\qquad a_s=\frac{y_{A,s}}{3},
\]
where $y_{H,s}\in\{0,1,2,3\}$ is the number of human-authored poems labeled ``human'' and $y_{A,s}\in\{0,1,2,3\}$ is the number of model-generated poems labeled ``human''. We analyzed the within-subject difference
\[
\delta_s=h_s-a_s.
\]
Across $N=50$ participants, the mean difference was $\bar{\delta}=0.020$ (SD $=0.439$), with a 95\% confidence interval of $[-0.105,\,0.145]$. A paired $t$-test (equivalently, a one-sample $t$-test of $\delta_s$ against 0) found no evidence that the mean difference differed from zero ($t(49)=0.3225$, $p=0.7485$; 95\% CI $[-0.105,\,0.145]$), indicating near-identical ``human'' labeling rates for human- and model-authored poems within participants.

\subsubsection*{Exact nonparametric confidence interval for the paired difference}

To provide a distribution-free summary of the within-subject difference $\delta_s$, we computed the Hodges--Lehmann estimator, defined as the median of Walsh averages $w_{ij}=(\delta_i+\delta_j)/2$ for $1\le i\le j\le N$.\cite{hodges1963estimates} We obtained an exact 95\% confidence interval by inverting the Wilcoxon signed-rank test.\cite{wilcoxon1945individual} Because $\delta_s$ is discrete in our setting (multiples of $1/3$), ties in $|\delta_s|$ are common.

The Hodges--Lehmann estimate was $\hat{\theta}_{HL}=0.00$, with an exact 95\% confidence interval of $[-0.1667,\,0.1667]$ (based on $m=41$ nonzero paired differences). This provides a robust nonparametric confirmation that the typical within-subject difference in ``human'' labeling rates between human- and model-authored poems is centered near zero. We additionally validated this result via bootstrap resampling (10,000 resamples), obtaining the same confidence interval endpoints.

\subsubsection*{Subject-level accuracy versus chance}

We also analyzed participants' discrimination performance at the subject level. For each subject $s$, an accuracy score was computed as
\[
\hat p^{(\text{acc})}_s=\frac{\text{number of correctly classified poems}}{\text{number of poems answered}},
\]
where a response was counted as correct if the participant's judgment matched the ground truth (AI vs.\ human). We had $50$ participants, each classifying $6$ poems. Across participants, accuracy was near chance (mean $=0.510$, SD $=0.219$, median $=0.500$).

\begin{figure}[h!]
    \centering
    \includegraphics[width=0.6\textwidth]{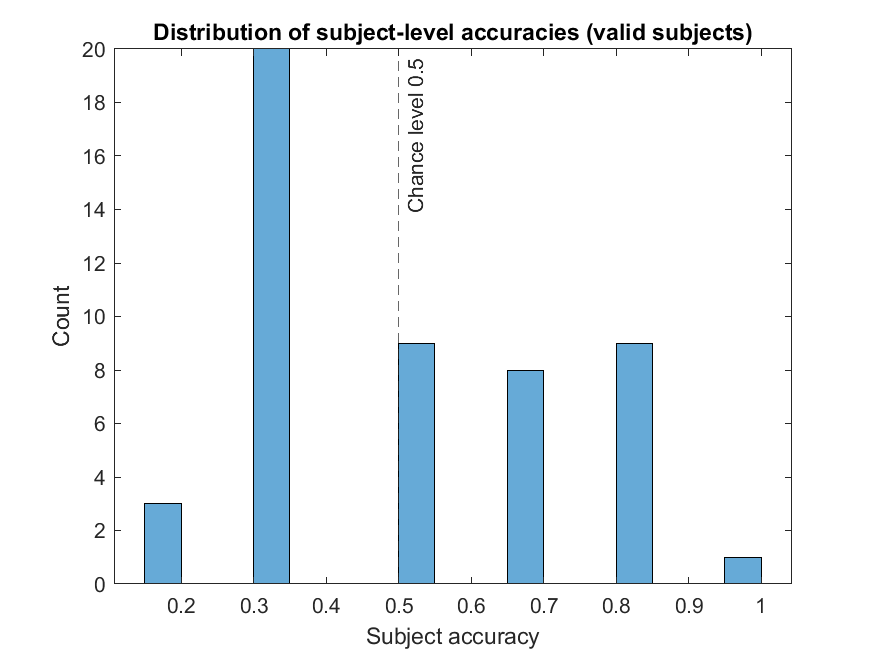}
    \caption{Distribution of subject-level accuracies across the 50 valid subjects. The dashed vertical line marks chance-level accuracy (0.5).}
    \label{fig:acc_hist}
\end{figure}

To test whether mean accuracy differed from chance (0.5), we conducted a one-sample $t$-test of $H_0:\mu_{\text{acc}}=0.5$ versus $H_1:\mu_{\text{acc}}\neq 0.5$. The test yielded $t(49)=0.322$, $p=0.7485$, with a 95\% confidence interval for the mean accuracy of $[0.448,\,0.572]$. We also performed a Wilcoxon signed-rank test on the subject accuracies, testing the null hypothesis that the median accuracy equals 0.5; the test yielded $W=467.5$, $p=0.6166$.\cite{wilcoxon1945individual} Figure~\ref{fig:acc_hist} shows the distribution of subject-level accuracies, with a vertical reference line at 0.5 corresponding to chance-level performance.

\subsection*{Analysis of the Poetic Corpus of Naomi Efron}

Linguistic and Semantic Foundations Analysis of the linguistic corpus of the poetry collection reveals a clear dominance of semantic fields related to dimensions of time, acoustic states, and acts of documentation and preservation. 
The central organizing principle is the concept of "time," which appears with high frequency through terms such as "hour," "moment," and "clock," functioning as either a dynamic entity or a fixed structural framework. Alongside the temporal axis, there is a significant quantitative presence of terms describing acoustic states of silence and stillness, which are defined as both the subject and the object of the cognitive actions of listening and reception. The speaker's identity and the poetic process are anchored in representational systems including the terms "word," "poem," "page," and "name," where the concept of "name" serves as a pivot for examining processes of commemoration, erasure, and self-definition.
As to the verbal and spatial dynamics at the verbal level, the text is propelled by a system of verbs expressing cognitive actions and investigative processes, primarily "to write," "to remember," and "to ask". The act of writing is consistently defined as a tool for preserving information and a means of coping with states of lack. 
Furthermore, there is a prominent presence of verbs denoting stability and persistence, such as "remained" and "left over," which appear in the context of physical traces and memories. Spatial perception is largely represented through the preposition "between," appearing with unusual frequency to establish liminal states and conceptual boundaries between opposites, such as clouds and dust, sirens and breathing, or memory and forgetfulness.
Metaphorical structure and stylistic definitions' analysis reveals a pattern of intersecting abstract concepts with physiological systems or rigid architectural structures. This includes connecting the "heart" to architectural elements like a floor or a window and treating "memory" as a tangible storage space represented by drawers and bags. Subjective experiences and emotions are transformed into objects with physical presence and weight through material metaphors such as "cup," "bread," or "fuel". These findings justify defining Naomi Efron’s linguistic system as a distinct "poetic style" characterized by structural and rhetorical consistency.
The poetic style crystallizes through the tension between abstract themes and a consistent tendency toward the "concretization of the abstract". Turning memory or silence into objects with volume and location, such as a "bag of silences" or a "drawer of years", is a primary stylistic hallmark. 
Another central element is syntactic liminality, created by the intensive use of the word "between," which defines reality as a space of mediation and permanent uncertainty. Rather than describing absolute states, Naomi Efron focuses on the gap between opposites, lending the work a reflective and investigative tone.
Efron’s poetry maintains a consistent epistemological stance, favoring verbs of reception and preservation (to listen, to remember, to remain). This style does not seek an active description of events but rather adopts the position of a "witness" observing the remnants and traces left behind. 
Consequently, language becomes a tool for self-poetic inquiry, where the poems function as both the subject and the instrument. In summary, the statistical consistency of key concepts, the structural use of material metaphors, and the reliance on a syntax of mediation confirm the existence of a cohesive and unique poetic style.
\subsubsection*{Corpus Size and Split}

The total corpus comprises 2508 words.

The text is divided into two primary functional segments:
\begin{itemize}
  \item \textbf{Introductory/Theoretical Segment:} Includes the Prologue and the Poetic Manifest, which establishes the ontological and methodological framework of the work (220 words).
  \item \textbf{Poetic Body:} Includes 50 poems, some of which are grouped into 3 poem cycles (2288 words).
\end{itemize}
\subsubsection*{Top Anchor Semantic Fields}
\begin{table}[tbp]
\caption{Leading semantic fields}
\label{tab:semantic_fields}
\centering
\begin{tabular}{|l|l|r|}
\hline
\textbf{Semantic Domain} & \textbf{Lexical fields: examples} & \textbf{Word Counting} \\
\hline
Identification & Name, Naming & 12 \\
Subjectivity & I, Me & 33 \\
Ars-Poetics & Poem, Poetry, Writing & 31 \\
Physical Terms: Optics & Light, Dark, Shadow & 25 \\
Physical Terms: Acoustics & Voice, Silence, Sound & 61 \\
Physical Terms: Time & Time, Year, Clock & 37 \\
Emotion & Heart, Feeling, Imagination, Love, Fear & 37 \\
Movement in space & Remain, Stay, Go, Walk & 36 \\
Domestic Spaces & House, Floor, Ceiling, Window, Door, Drawer, wall & 23 \\
Perception & See, Hear, Whisper, Talk & 19 \\
Nature & Tree, Stone, Grass, Sun, Sea & 11 \\
\hline
Total &  & 325 \\
\hline
\end{tabular}
\end{table}

Table \ref{tab:semantic_fields}  presents the main semantic fields distribution. The table reveals a clear dominance of acoustic, temporal, and affective domains, suggesting an inward, voice-centered poetics. Physical space and natural imagery remain secondary, functioning primarily as a minimal stage for subjective and meta-poetic articulation. The corpus privileges sound, time, and interiority over visual or material description.

\end{document}